\documentclass[twoside,11pt]{article}

\usepackage{automl2020}
\usepackage{amsmath,cases}
\usepackage{subcaption}
\usepackage{wrapfig}
\usepackage{sidecap}
\usepackage{xcolor}

\jmlrheading{S. Glass, S. Spasov and P. Li\`o}

\ShortHeadings{RicciNets}{Glass, Spasov and Li\`o}
\firstpageno{1}

\begin{document}

\title{RicciNets: Curvature-guided Pruning of High-performance Neural Networks Using Ricci Flow}

\author{\name S. Glass \email seg67@cam.ac.uk \\
       \addr Department of Physics, University of Cambridge
       \AND
       \name S. Spasov \email ses88@cam.ac.uk \\
       \addr Department of Computer Science and Technology, University of Cambridge
       \AND
       \name P. Li\`o \email pl219@cam.ac.uk \\
       \addr Department of Computer Science and Technology, University of Cambridge}

\maketitle

\begin{abstract}
A novel method to identify salient computational paths within randomly wired neural networks \textit{before} training is proposed. The computational graph is pruned based on a node mass probability function defined by local graph measures and weighted by hyperparameters produced by a reinforcement learning-based controller neural network. We use the definition of Ricci curvature to remove edges of low importance before mapping the computational graph to a neural network. We show a reduction of almost 35\% in the number of floating-point operations (FLOPs) per pass, with no degradation in performance. Further, our method can successfully regularize randomly wired neural networks based on purely structural properties, and also find that the favourable characteristics identified in one network generalise to other networks. The method produces networks with better performance under similar compression to those pruned by lowest-magnitude weights. To our best knowledge, this is the first work on pruning randomly wired neural networks, as well as the first to utilize the topological measure of Ricci curvature in the pruning mechanism.
\end{abstract}

\section{Introduction}
At birth, the construction of the most important networks is largely random and random graph modelling is heavily used in the study of the human brain \citep{RandomGraph, RandomGraph2}. Recent work on randomly wired neural networks has emulated this in the field of deep learning, and moves away from the wiring approach that has typically dominated NAS \citep{RandWire}. Randomly wired networks display comparable performance to state-of-the-art architectures (eg. ResNet, DenseNet), and provide a relatively unrestricted space on which to perform further optimisation. We propose a search method that takes place within a low-dimensional search space; a pruning methodology which operates on networks produced by a successful random network generator. It is based on the discrete Ricci curvature of a graph, with estimates of a node's community, contribution to robustness and computational demand contributing to the identification of salient computational paths in the network. A curvature-guided diffusion process, Ricci flow, deforms the discrete space of the graph, and edges within the graph are removed based on their local deformation. A reinforcement learning controller parameterises the Ricci flow process. To the best of the authors' knowledge, this is the first work to take inspiration from the physics concept of space curvature deformation, in combination with reinforcement learning, to drive the process of neural network pruning.

The technique proposed has three key advantages: (1) it is a successful form of regularisation which promotes sparse network connectivity, and as a result low computational demand; (2) it operates with no degradation in baseline performance; (3) a successful hyperparameter state can be applied, without further optimisation, to similarly produced random networks. The process operates before training, saving compute during both training and inference. This is the first known work to investigate the pruning of randomly wired neural networks. Using \textit{RicciNets}, we demonstrate novel, efficient generation of \textit{compact} neural architectures.
\section{Related Work}
\label{Related Work}
Ricci curvature is the definition of curvature used in Einstein's Field Equations. Loosely, it measures the deformation of a volume on the surface of a manifold relative to the volume in Euclidean space (see Appendix A).
Unlike other ML works, in which continuous Ricci curvature is used to visualise high-dimensional loss landscapes \citep{LossLandscapes} or in novel characterisations of structural features \citep{TDA, NeuralPersistence}, we use a measure of Ricci curvature within the discrete space of a graph. In order to do this, we relate Ricci curvature to optimal transport \citep{OllivierRicci}. Given a probability measure at each node, optimal transport can be formulated on a graph and Ricci curvature can be calculated. For a metric space $(X,d)$ equipped with probability measure $m_x$ for each $x \in X$, the Ollivier-Ricci curvature, $\kappa$, along the shortest path \textit{xy} is given by Eq. (\ref{ricciCurvature}), where $W(m_{x},m_{y})$ is the Wasserstein distance, and $d(x,y)$ is the path distance:
\begin{equation}
\kappa(x,y) = 1 - \frac{W(m_{x},m_{y})}{d(x,y)}.
\label{ricciCurvature}
\end{equation}
Similar to \citet{CommDetection}, we use a curvature-guided diffusion process, Ricci flow, to detect community structures within a network. The probability distribution used here includes further terms to estimate the computational demand of an individual node as well as its contribution to the overall network’s robustness with respect to damage. In both works, the curvature evolves under discrete time intervals, Eq. (\ref{discreteTime}).
\begin{equation}
w_{ij}^{k+1} = (1 - \kappa_{ij}^{(k)})d^{(k)}(i,j).
\label{discreteTime}
\end{equation}
Successful efforts in NAS may require months or even years of compute time. \citet{ZophNAS} demonstrated a computationally expensive process in which a RL controller parameterised a search within a high-dimensional search space. Resultant architectures match state-of-the-art performance. We optimise a combination of three hyperparameters for our search. Randomly wired networks offer a relatively unrestricted initial search space with good baseline performance. \citet{RandWire} found a Watts-Strogatz graph generator \citep{watts1998collective} produced networks with the best performance, with $WS(K = 4,P = 0.75)$ and $N = 32$ (see Appendix B). Their simple graph-to-network mapping allowed a focus on wiring and structural features. The successful generator and straightforward mapping are both used here.
\section{Method}
\label{Method}
Random computational graphs are generated using the Watts-Strogatz model. First, we use a controller neural network to predict a hyperparameter state. Second, we propose a node mass distribution based on local graph measures weighted by the predicted hyperparameters. Then, we use the process of Ricci flows to compute weights associated with each edge and prune the computational graph based on an edge threshold value. The pruned computational graph is mapped to a neural network and trained on the dataset. Accuracy and FLOPs per pass are combined to yield a reward then passed to the controller network, which is updated via a policy gradient method (see Appendix C). The code is publicly available at https://github.com/seglass5/RicciNets.
\subsection{Mass Distribution}
We calculate the curvature within a network using a hypothesised probability (mass) distribution as  Eq. (\ref{massDist}):
\begin{equation}
\label{massDist}
m_{x}^{\alpha,\beta,\gamma,\delta}(x_i)=\begin{cases}
  \alpha & \text{for $x = x_i$} \\
  (1 - \alpha) \bigl[\beta (\frac{1}{\textup{Deg(x)}}) + \gamma (\textup{Input(x)}) + \delta (\frac{\textup{Output(x)}}{\textup{Input(x)}})\bigr] & \text{for $x \in \pi(x)$} \\
  0 & \text{Otherwise}.
  \end{cases}
\end{equation}
Input($x$) defines the input degree of node $x$, Output($x$) the output degree and Deg($x$) the total degree. $\pi(x)$ defines the immediate neighbours of $x$. $\alpha$ gives the proportion of mass to remain on a node. $\beta$, $\gamma$ and $\delta$ control the contribution of each of the three terms. The first term promotes a well-defined community structure, $\frac{1}{\textup{Deg(x)}}$ yields a lower mass for a node with more neighbours. The second term promotes low input degree. Taking the transformation operation at a node to be of linear complexity in input degree, this approximates to promoting low computational burden at each node. The final term promotes a smaller ratio of output degree to input degree. A greater loss in accuracy is observed for removal of a node with high output degree, and an edge with low target node input degree \citep{RandWire}. A smaller ratio of output degree to input degree is therefore taken to indicate better robustness with respect to graph damage. By requiring that the masses of a node and its neighbours sum to unity, this can be reduced to an equation in three hyperparameters.
\subsection{Pruning Threshold}
The weight associated to each edge in the graph is updated via Ricci flow from an initial value of zero, Eq. (\ref{discreteTime}). The weights are normalised to prevent expansion to infinity and checked for convergence on each iteration. The process of Ricci flow ran for 50 iterations and typically reached convergence well within this limit. The threshold for pruning was set to the mean of all the weights in the network following Ricci flow. Hyperparameter selection can alter the skewness of the distributions of curvatures and weights, and adjusting the distribution of weights is interpreted as the controller network learning a definition of saliency; if very few paths can be considered salient, parameter prediction can lead to negative skewness in the weight distribution and more edges are removed. Similarly, if the drop in accuracy is too high for removing a group of paths, a set of hyperparameters that adjusts the mean to save this group can be learnt.
\subsection{Controller}
An auxiliary controller network generates hyperparameter states. The controller is implemented as a simple feed forward network. The parameter states are one-hot encoded and discretised in the range $[0,1]$. The output parameters, $[\alpha, \beta, \delta]$ are passed to the pruning function.  The controller seeks to maximise its expected reward, Eq. (\ref{REINFORCERewards}), where $J(\theta_c)$ indicates the expected reward at a parameter state $\theta_c$. $a_{1:T}$ represents the list of actions (possible hyperparameter combinations) for $T$ hyperparameters,
\begin{equation}
J(\theta_c) = E_{P(a_{1:T}; \theta_c)}[R].
\label{REINFORCERewards}
\end{equation}
Since the reward signal is non-differentiable, we use a policy gradient method to iteratively update $\theta_c$. We use the REINFORCE rule \citep{Williams1992}, Eq. (\ref{REINFORCE}).
\begin{equation}
_{\nabla \theta_c}J(\theta_c) = \sum_{t = 1}^{T} E_{P(a_{1:T}; \theta_c)} [_{\nabla \theta_c}log(P(a_t | a_{(t-1):1};\theta_c))R].
\label{REINFORCE}
\end{equation}
An empirical approximation of the above quantity is given in Eq. (\ref{empiricalREINFORCE}). $m$ is the number of parameter states sampled in one batch by the controller. The reward that the network in the $k^{th}$ parameter state achieves after training is $R_k$,
\begin{equation}
\frac{1}{m} \sum_{k = 1}^{m} \sum_{t = 1}^{T} [_{\nabla \theta_c}log(P(a_t | a_{(t-1):1};\theta_c))R_k].
\label{empiricalREINFORCE}
\end{equation}
The reward used to update policy is given in Eq. (\ref{reward}). The top one accuracy, $A$, is regularised by the FLOPs per pass of the network $F$ using a regularisation parameter $\mu$.
\begin{equation}
J(\theta_c) = A - \mu \frac{F}{F_{baseline}}.
\label{reward}
\end{equation}
Episode rewards are discounted according to Eq. (\ref{discount}), where $v(\theta_c)$ is the episode reward for a state $\theta_c$, $\gamma$ a discount parameter, and $k$ the number of iterations within an episode. This encourages prolonged episodes.
\begin{equation}
v(\theta_c) = \sum^{N}_{k = 0} \gamma^{k} J(\theta_{c+k}).
\label{discount}
\end{equation}
\section{Experiments and Results}
\label{Experiments}
Experimentation is based on the classification of images from the CIFAR-10 dataset, with 50,000 training images and 10,000 test images. The images are batched in groups of 64. Each network is trained for 4 epochs, and the policy gradient controller ran over 20 episodes, with episodes batched in pairs to update policy.
\subsection{Evaluation Procedure}
Network performance is evaluated using the top-one accuracy and the number of FLOPs per pass of the network produced. Performance is measured in relation to a baseline set by an unpruned network produced using the same generator parameters. To assess the importance of considering the topology of randomly wired networks in the pruning procedure, we compare \textit{RicciNets} against pruning weights by lowest magnitude \citep{46512}. The combination of hyperparameters learnt for a given graph is applied to other graphs of the same random graph generator. We note that while this implementation of randomly wired neural networks yielded an accuracy of $91.5 \pm 0.2 \%$ on CIFAR-10 over 100 training epochs, we only report results achieved after 4 epochs owing to resource constraints.
\subsection{Results}
\begin{figure}[h]
  \centering
    \begin{subfigure}[t]{0.4\textwidth}
        \centering
        \includegraphics[width=\textwidth,keepaspectratio]{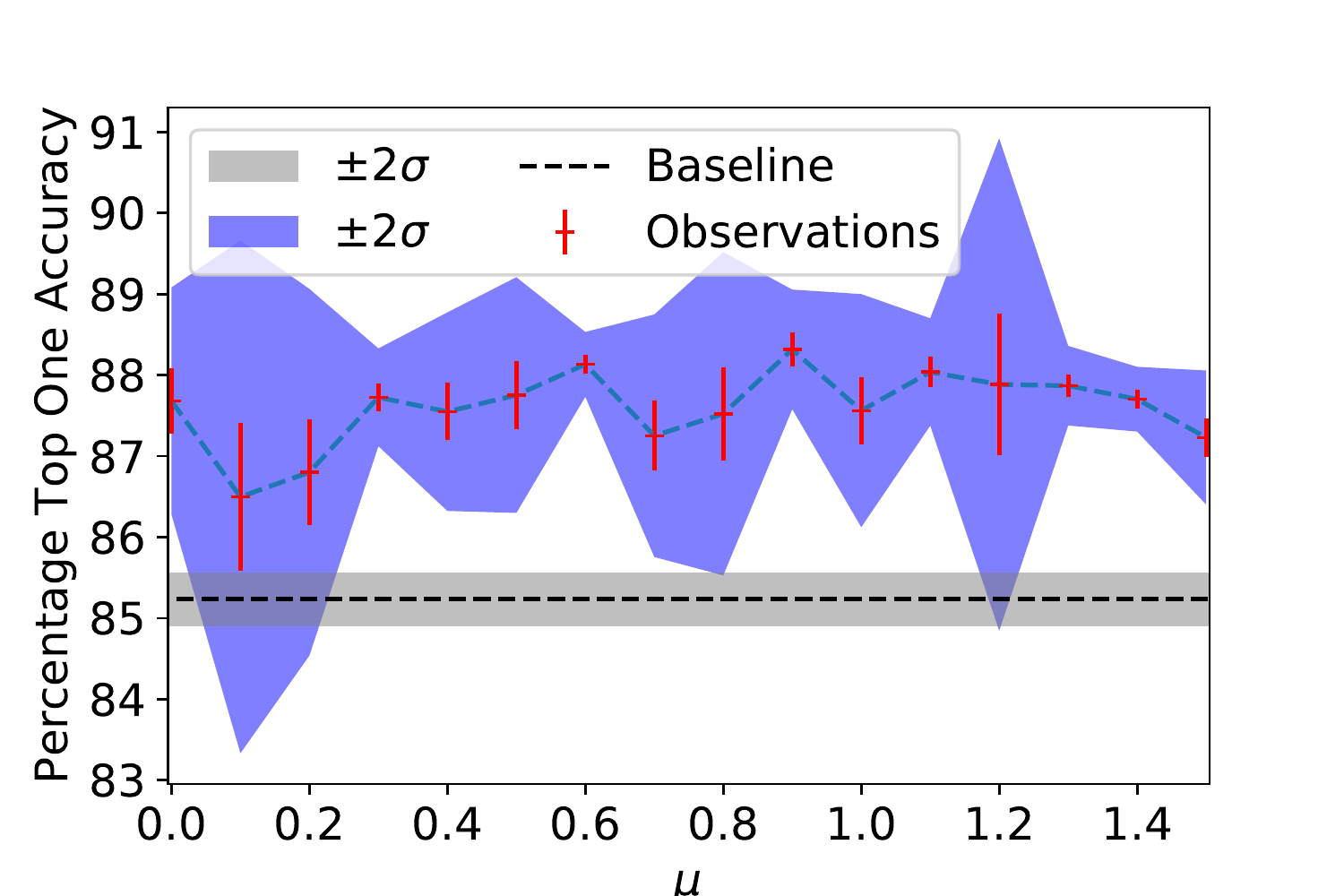}
        \caption{Accuracy against $\mu$. Pruned networks display better-than-baseline performance, with almost $3\%$ increase in accuracy on baseline when averaged across the range of $\mu$.}
        \label{fig:mutopone}
    \end{subfigure}%
    ~ 
    \begin{subfigure}[t]{0.4\textwidth}
        \centering
        \includegraphics[width=\textwidth,keepaspectratio]{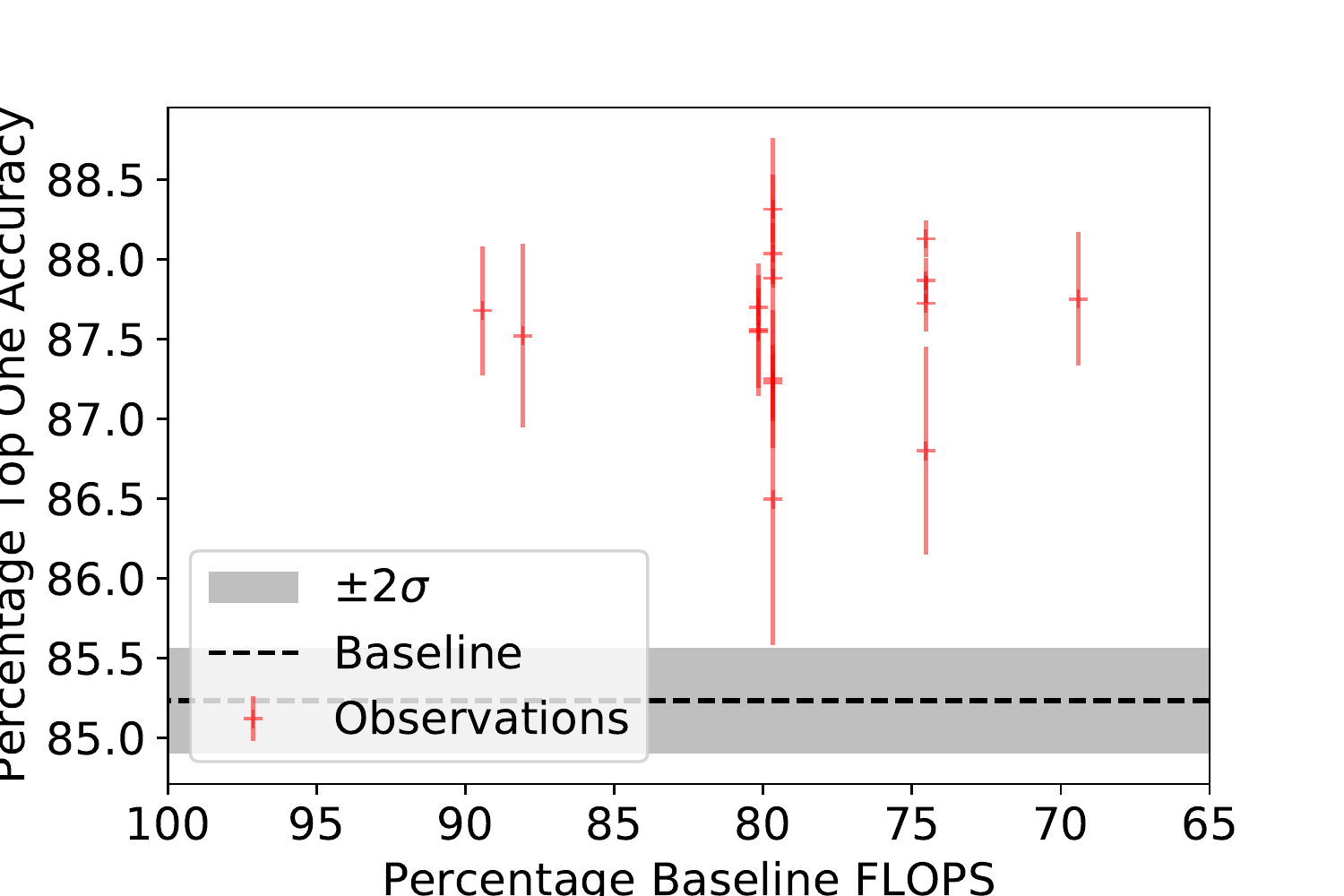}
        \caption{Accuracy gainst FLOPs per pass of pruned network as a percentage of baseline. Overlapping errorbars appear darker.}
    \end{subfigure}
    \caption{The pruned networks show no degradation in performance under compression.}
\label{fig: flops}
\vspace{-3.5mm}
\end{figure}

Fig. \ref{fig: flops} (a) shows the variation in top-one accuracy of the resultant networks for regularisation parameter $\mu$ in the range [0, 1.5]. The methodology produces better-than-baseline performance under compression across the range of $\mu$, with a small variance in top one accuracy ($\sigma^2 = 0.21$). (b) shows the top-one accuracy of architectures against the FLOPs per pass expressed as a percentage of baseline. All networks produced operated using $68 - 91 \%$ of the baseline FLOPs per pass. For small networks, more severe compression would result in a chain-like structure and a sharp drop off in accuracy, which would be discouraged by the controller (see Appendix D). An exploratory step carried out with probability $p$ within the controller, or further fine-tuning of the policy gradient network could result in a larger range of compression.
\textit{RicciNets} demonstrates a restricted range of compression when compared to pruning via lowest-magnitude weights. Within this range, however, the networks produced by \textit{RicciNets} demonstrate better performance than those pruned by weight, Table \ref{averages}. Future work includes incorporating more control over the level of compression via alternative ways to regularize the reward objective.

\begin{figure}[h]
  \centering
    \begin{subfigure}[t]{0.4\textwidth}
        \centering
        \includegraphics[width=\textwidth]{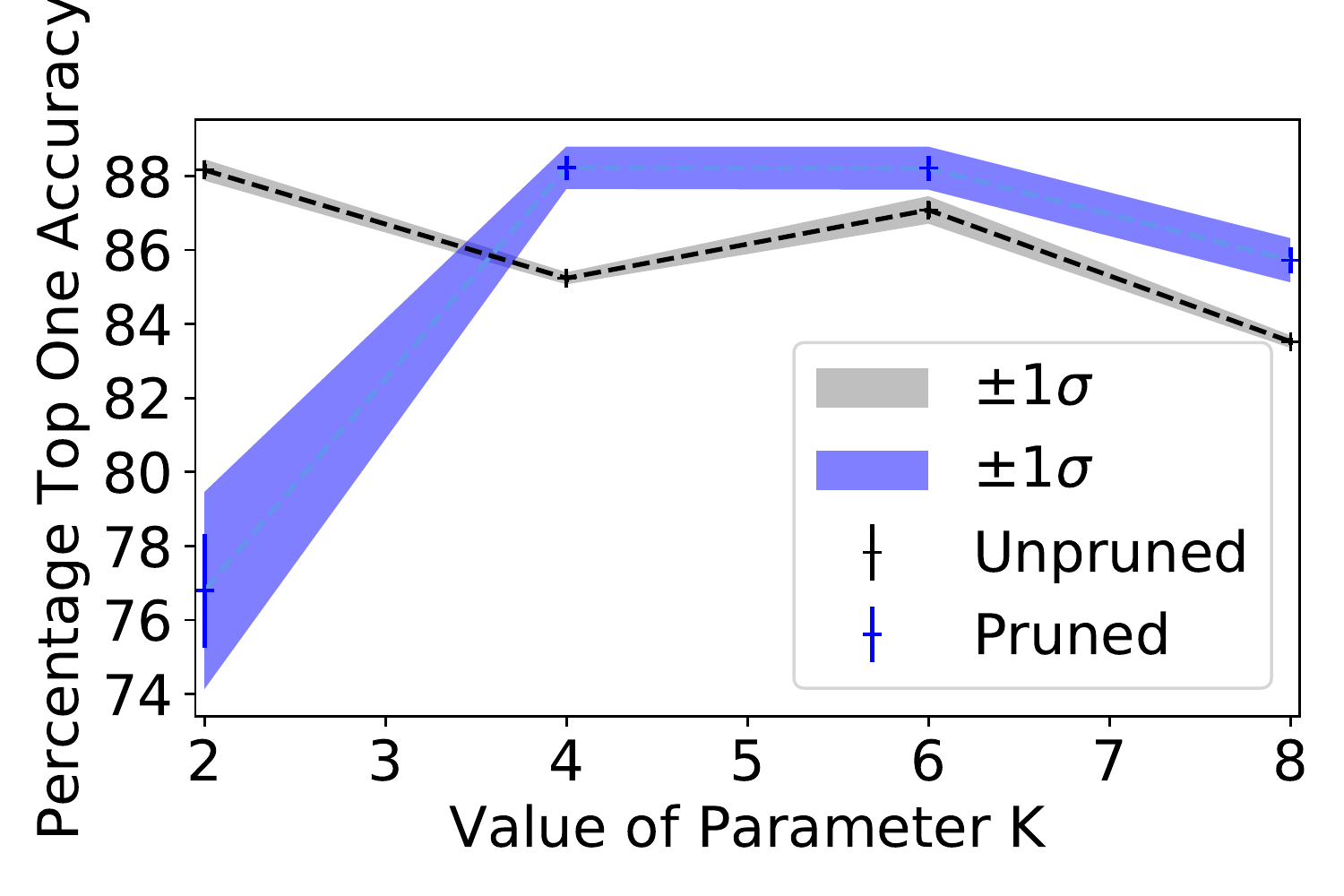}
        \caption{Top one accuracy against the value of parameter $K$ in $WS(K,P)$, with $P = 0.75$ and $N = 32$.}
    \end{subfigure}%
    ~ 
    \begin{subfigure}[t]{0.4\textwidth}
        \centering
        \includegraphics[width=\textwidth]{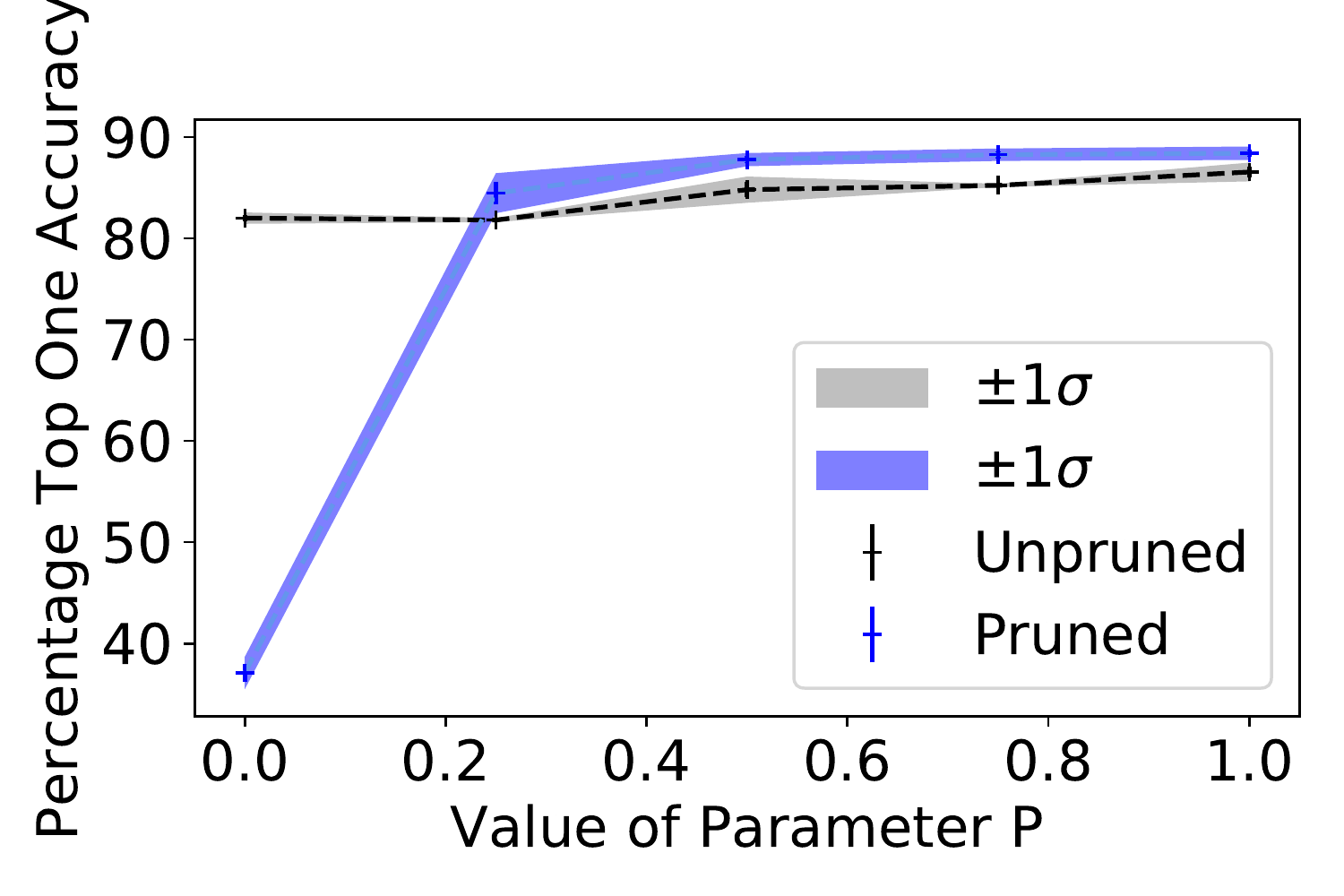}
        \caption{Top one accuracy against the value of parameter $P$ in $WS(K,P)$, with $K = 4$ and $N = 32$.}
    \end{subfigure}
    ~
    \begin{subfigure}[t]{0.4\textwidth}
        \centering
        \includegraphics[width=\textwidth,keepaspectratio]{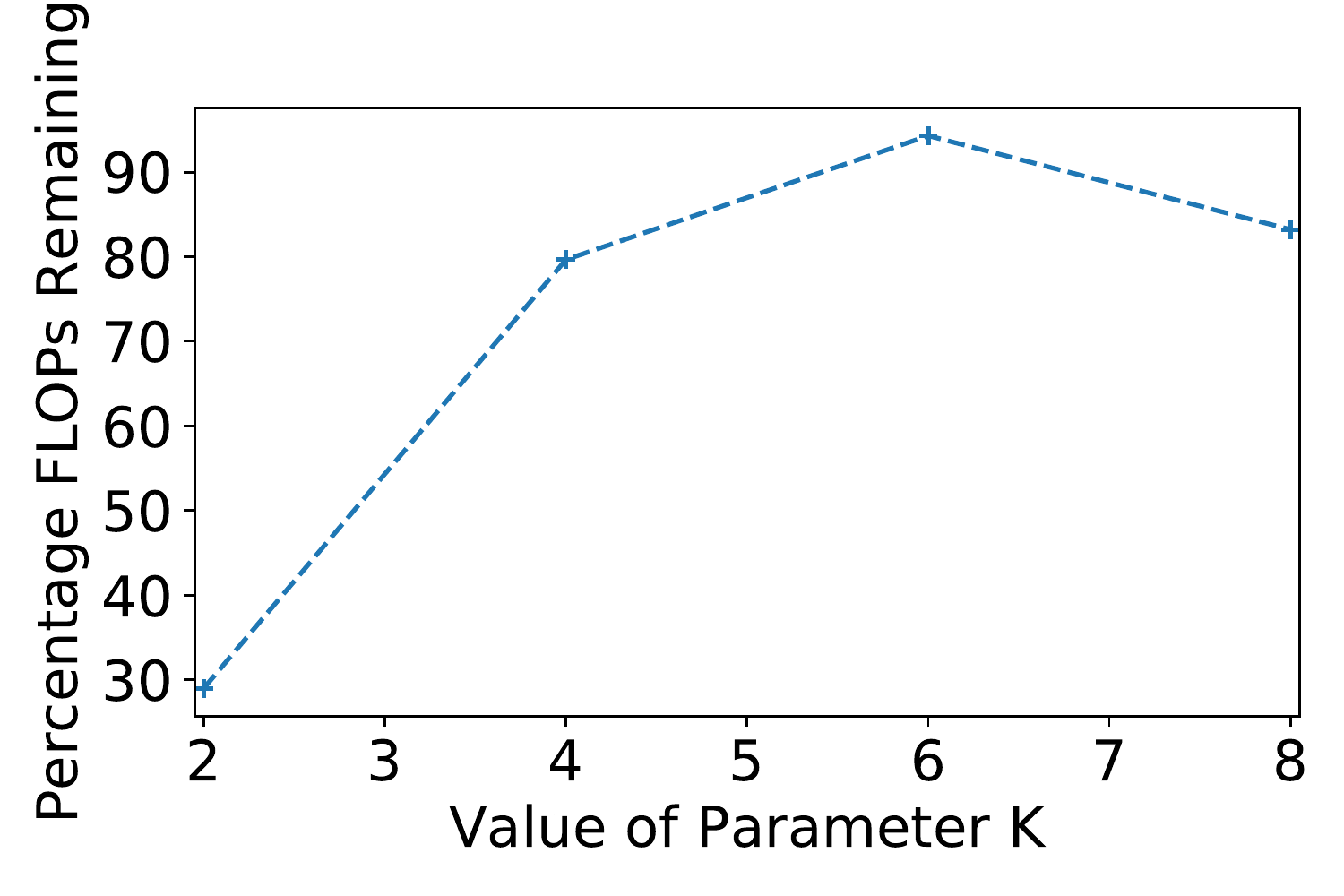}
        \caption{FLOPs per pass through the network against parameter value $K$ in $WS(K,P)$, with $P = 0.75$ and $N = 32$.}
    \end{subfigure}%
    ~ 
    \begin{subfigure}[t]{0.4\textwidth}
        \centering
        \includegraphics[width=\textwidth,keepaspectratio]{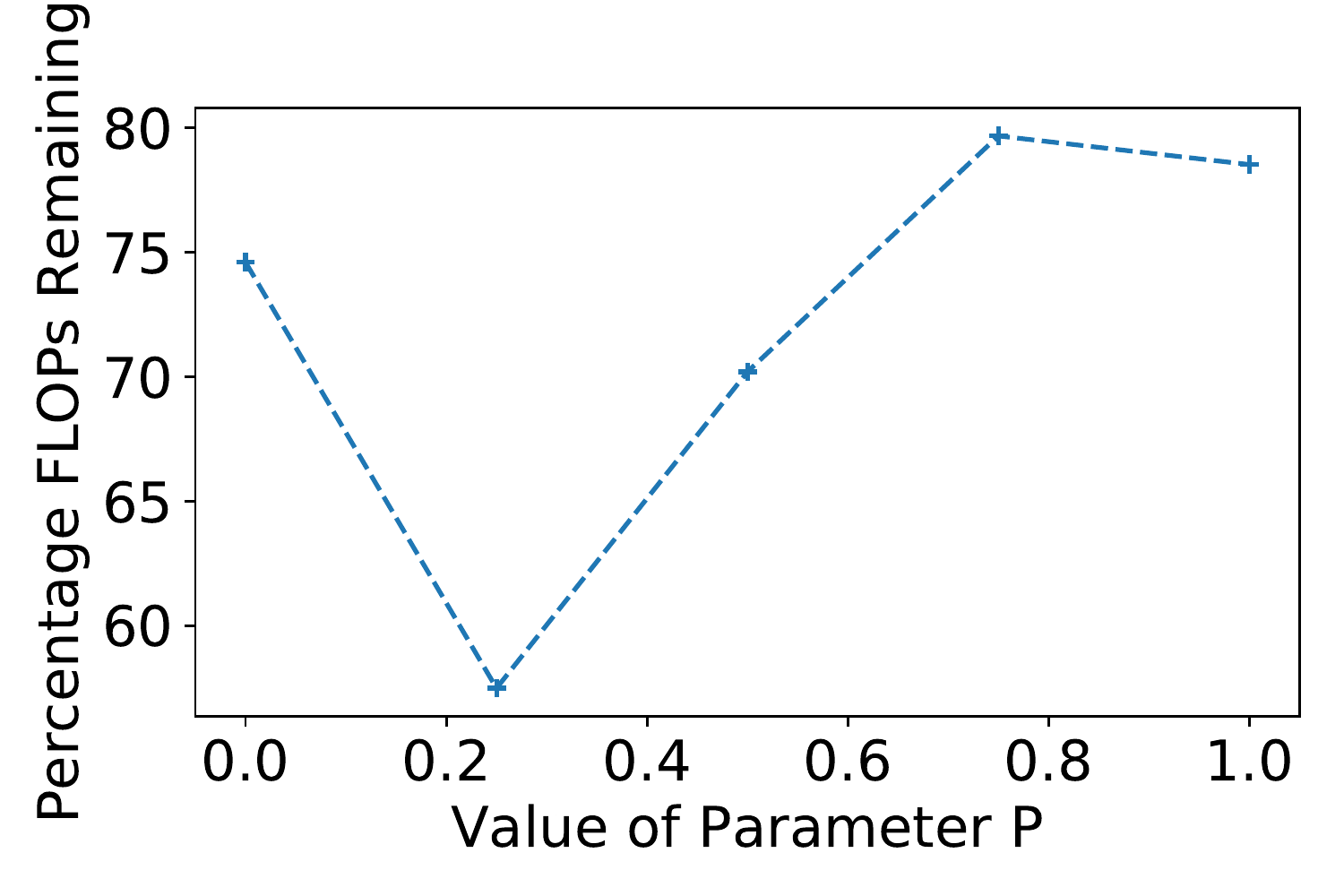}
        \caption{FLOPs per pass through the network against parameter value $P$ in $WS(K,P)$, with $K = 4$ and $N = 32$.}
    \end{subfigure}
    \caption{Pruning WS graphs without specific optimisation showed no drop in performance.}
\label{fig:noopt}
\centering
\vspace{-5mm}
\end{figure}

The combination of hyperparameters that produced the greatest accuracy using $WS(4, 0.75)$ generalised to other Watts-Strogatz graphs. Pruned networks generated with different $K$ and $P$ displayed an increase in performance and moderate compression, Fig. \ref{fig:noopt}. \textit{RicciNets} maintained the salient computational paths identified in the learnt case.
\begin{center}
\begin{table}[h]
\begin{center}

\caption{Average top one accuracy and average percentage baseline weights remaining after pruning for \textit{RicciNets}, pruning via lowest magnitude weights and baseline. Average taken within the $40 - 50 \%$ range of weights remaining.\label{averages}}
\begin{tabular}{ | c | c | c |  }

 \hline
 \textbf{Pruning}& \textbf{Top One Accuracy (\%)} & \textbf{Weights Remaining (\%)}\\
 \hline
 \textbf{RicciNets}   & $87.59 \pm 0.11$ & $41.90 \pm 0.47$\\
 \hline
 \textbf{Lowest Magnitude} &  $84.77 \pm 0.55$   & $45.00 \pm 3.54$\\
 \hline
 \textbf{Baseline} & $85.23 \pm 0.09$ & $100.00$\\
 \hline
\end{tabular}
\end{center}
\vspace{-4mm}
\end{table}
\end{center}

\section{Conclusions}
\label{Results}
Our model combines the principle of curvature with ML to carry out neural architecture search. It successfully identifies salient computational paths, and demonstrates a reduction in computational cost for no degradation in baseline performance. It outperforms pruning via lowest-magnitude weights on randomly wired neural networks. A combination of hyperparameters learnt with a given network generalises to others from the same generator with no specific optimisation, offering compression for no drop in performance. The results obtained suggest a successful novel methodology for compact NAS, and are the first on the compression dynamics of randomly wired neural networks. Future work will develop more comparative methods against other pruning procedures, and investigate off-policy controller algorithms.

\clearpage

\vskip 0.2in
\nocite{PhysRevE.101.062207}
\nocite{book:1020817}
\nocite{book:1035198}
\nocite{book:1146098}
\nocite{book:1581002}
\nocite{book:272733}
\nocite{molchanov2016pruning}
\nocite{MorphNet}
\bibliography{mybib}


\newpage

\appendix
\section{Ricci Curvature}

The notion of curvature was introduced by Gauss and Riemann over 190 years ago; it is a measure of how space is curved at a point in that space. Given an \textit{n}-dimensional manifold, which we define as a space that locally looks \textit{n}-dimensional, we may form a Riemannian metric; this assigns each tangent space of the manifold a Euclidean metric, which in turn gives the "standard" distance between any two vectors in the space. A manifold together with its corresponding Riemannian metrics forms a Riemannian manifold \citep{CommDetection}.

For a surface \textit{S}, the Gaussian map from \textit{S} to the unit sphere sends a point on \textit{S} to the unit normal vector of \textit{S} at \textit{p}, a point on the unit sphere. The Gaussian curvature of the surface at a point \textit{p} is the Jacobian of the Gaussian map at \textit{p}, the signed area distortion of the Gaussian map at \textit{p}. Hence, the plane has zero curvature, the sphere has positive curvature, and the hyperboloid of one sheet has negative curvature. The curvature depends only on the induced Riemannian metric on the surface and does not depend on how the surface is embedded in space. 

Riemann generalised Gaussian curvature to higher dimensions. For a Riemannian manifold \textit{(M,g)}, the sectional curvature assigns each 2-dimensional linear subspace \textit{P} in the tangent space of \textit{M} at \textit{p} a scalar, the Riemannian sectional curvature. The scalar is equal to the curvature of the image of \textit{P} under the exponential map. A positively curved space tends to have small diameter and is geometrically crowded; a sphere, for example. Conversely, a negatively curved space is geometrically spreading out.

The Ricci curvature assigns each unit tangent vector \textit{v} at a point \textit{p} a scalar which is the average of the sectional curvatures of planes containing \textit{v}.

There have been various approaches to generalize the concept of curvature to non-manifold spaces. Here, we look to assign curvature to a graph, \textit{G(V, E, w)}, with vertices \textit{V}, edges \textit{E} and edge weights \textit{w}. Ollivier-Ricci curvature \cite{OllivierRicci} relates Ricci curvature to optimal transport, allowing a mapping to discrete spaces. Given a probability measure at each point, optimal transport can be formulated on general metric spaces and may be used to define Ricci curvature on a network with edge weights and probability measures at each vertex.

\section{Watts-Strogatz Random Graph Generator}

The Watts-Strogatz method demonstrated the most success within \cite{RandWire}. This operates by first placing $N$ nodes regularly in a ring, with each node connected to its $K/2$ neighbours on both sides, where $K$ is an even number. Then, in a clockwise loop, for every node $v$, the edge that connects $v$ to its clockwise $i^{th}$ next node is rewired with probability $P$. "Rewiring" is defined as uniformly choosing a random node that is not $v$ and that is not a duplicate edge. This loop is repeated $K/2$ times for $1 \leq i \leq K/2$. $K$ and $P$ are the only two parameters of the Watts-Strogatz model. Any graph generated by a state $(K, P)$ has exactly $N \cdot K$ edges. $(K, P)$ only covers a small subset of all possible $N$-node graphs, and a different subset than that covered by other random graph generators with equal $N$. Random graph generators present a relatively unrestricted initial search space, but a prior is introduced in the choice of random graph generator. Watts-Strogatz graphs display small world properties; the typical distance between randomly chosen nodes is proportional to $log(N)$.

\section{Method Overview}
\begin{figure}[h]
\includegraphics[width=\textwidth,height=0.8\textheight,keepaspectratio]{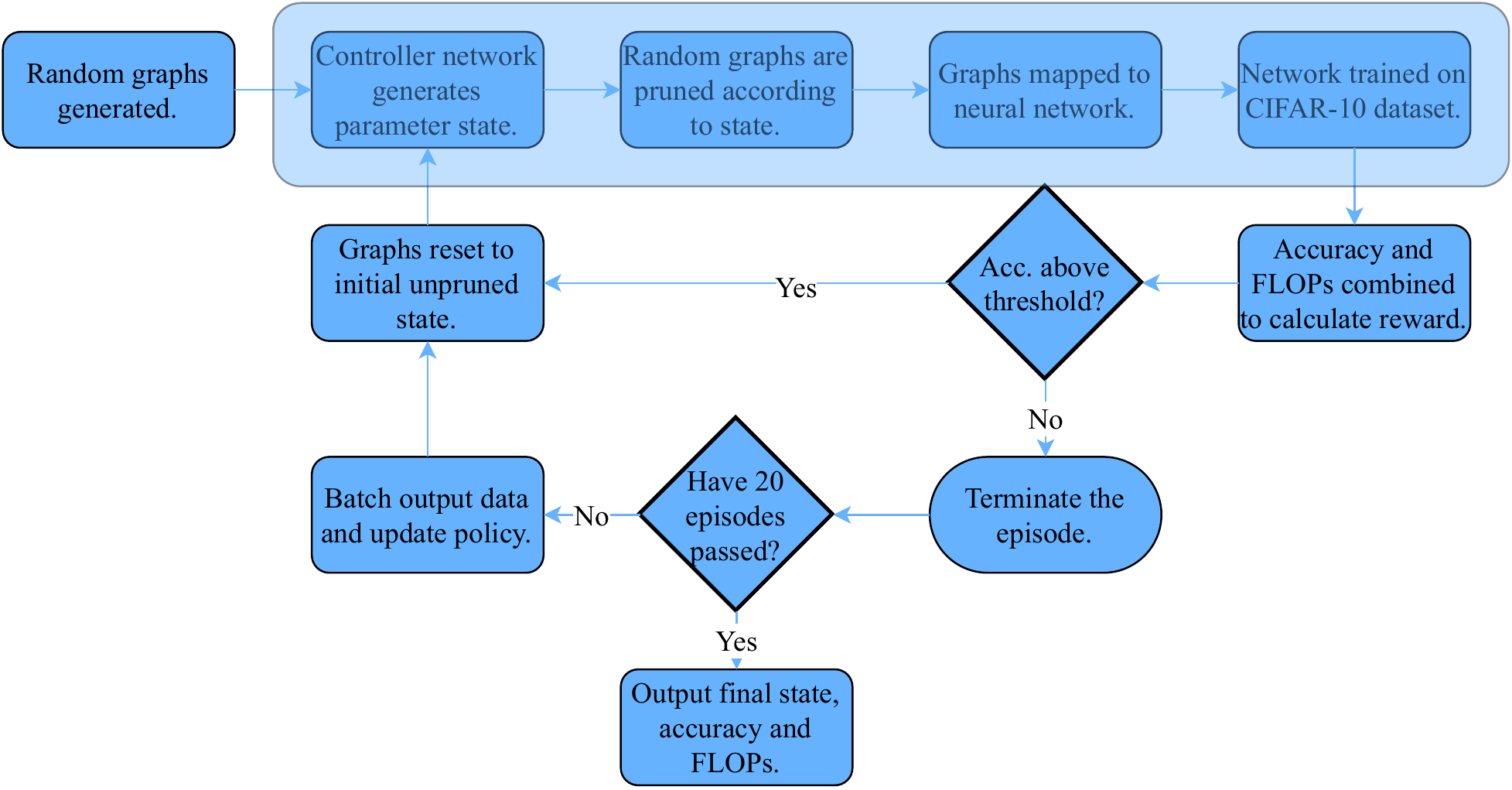}
\caption{An overview of the methodology. Pruning takes place before the graph-to-network mapping. The pale blue box indicates a single step within a policy gradient episode.}
\end{figure}
\clearpage
\section{Extensive Pruning}

\begin{figure}[h]
    \centering
    \begin{subfigure}[t]{0.5\textwidth}
        \centering
        \includegraphics[height=2.1in]{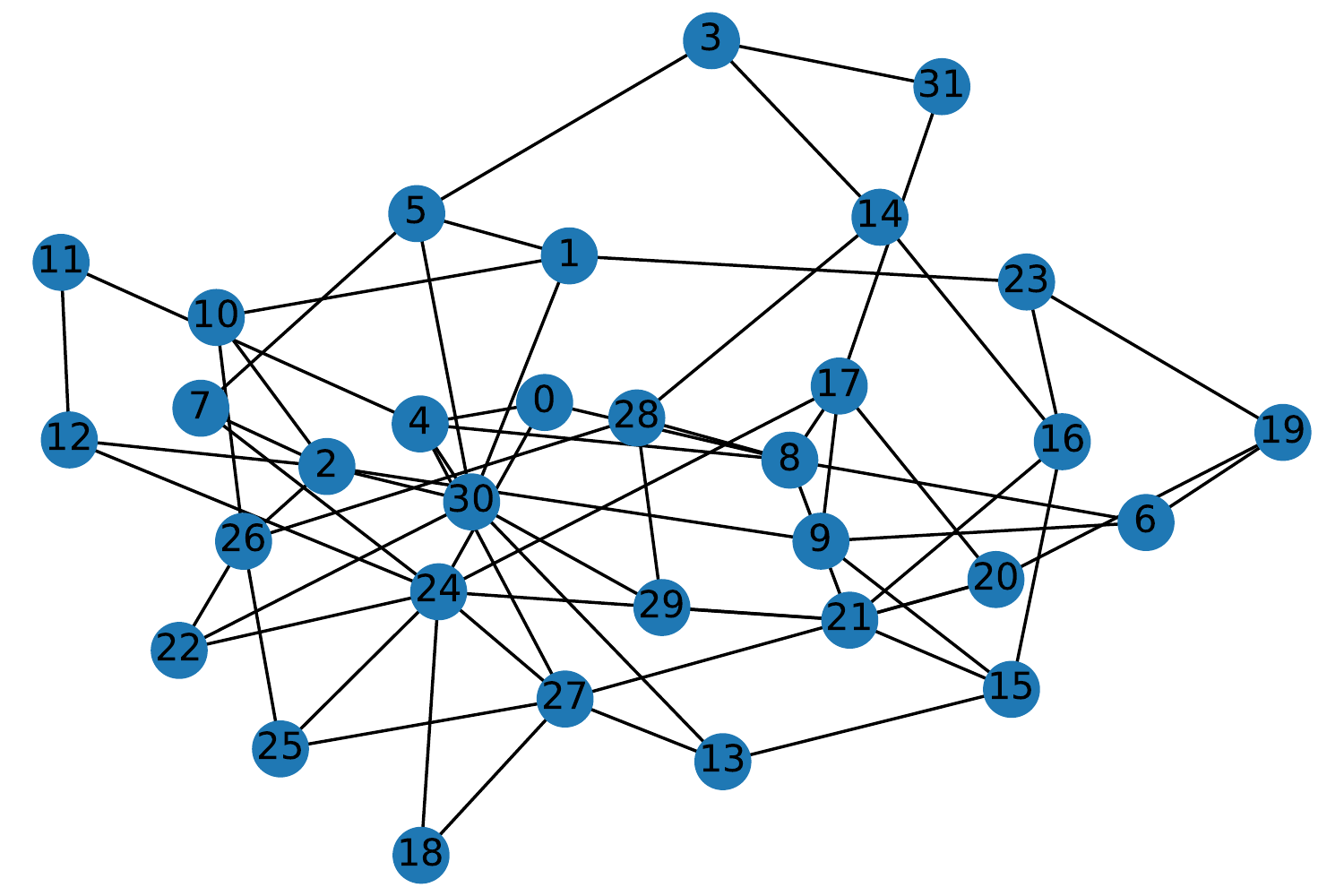}
    \end{subfigure}%
    ~ 
    \begin{subfigure}[t]{0.5\textwidth}
        \centering
        \includegraphics[height=2.0in]{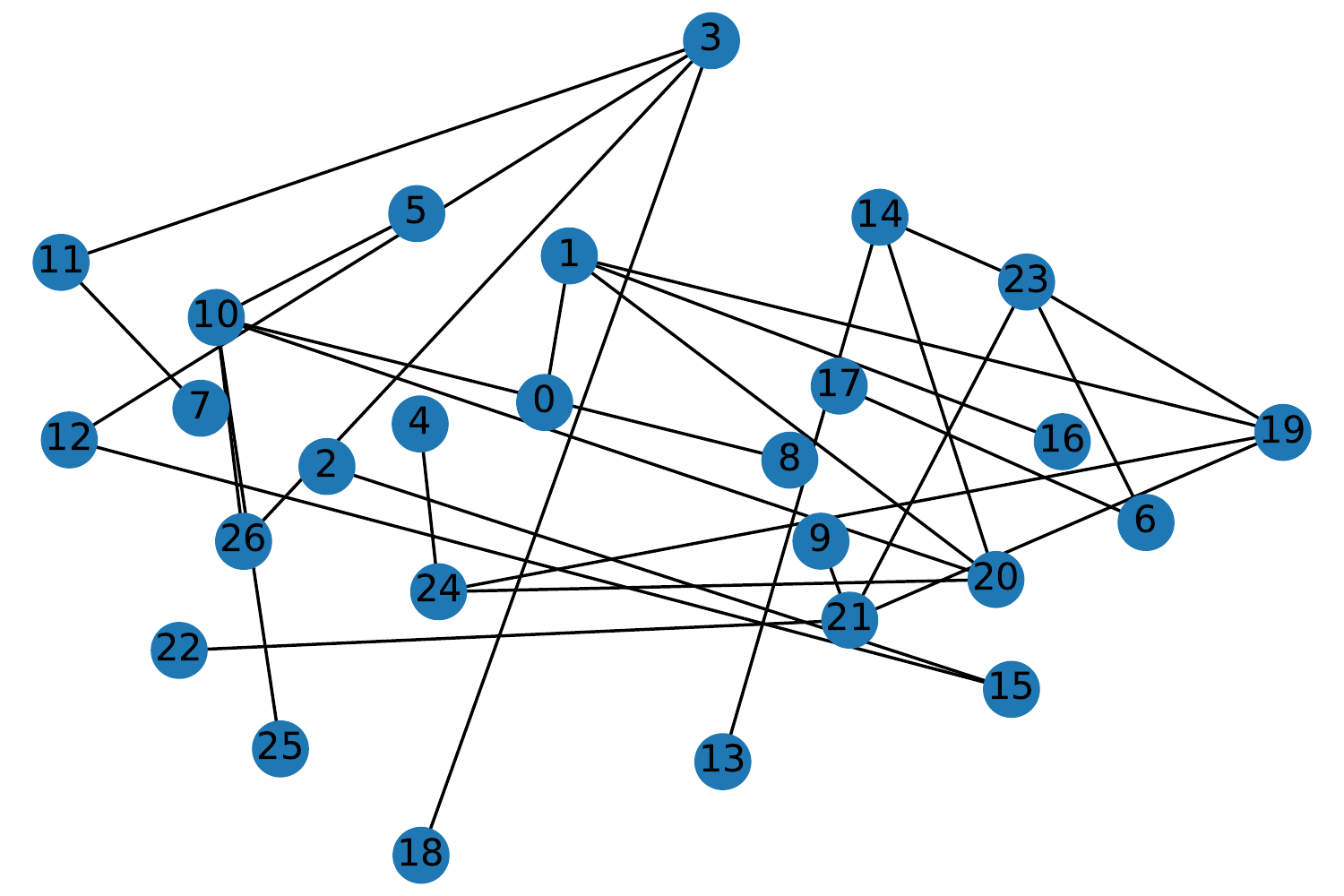}
    \end{subfigure}
    \caption{The unpruned state of a graph, $WS(4,0.75)$ with $N = 32$, left, and the same graph following pruning under a selected combination of hyperparameters, right. We observe increased sparsity whilst still retaining some clustering and skipped connections. Since both nodes and paths are removed, the node labels do not carry over from the unpruned to the pruned state, and are shown here for the purposes of information flow; in both cases data is carried in the direction of increasing node label.}
\label{fig:graphs}
\end{figure}

\includegraphics[width=0.9\linewidth,keepaspectratio]{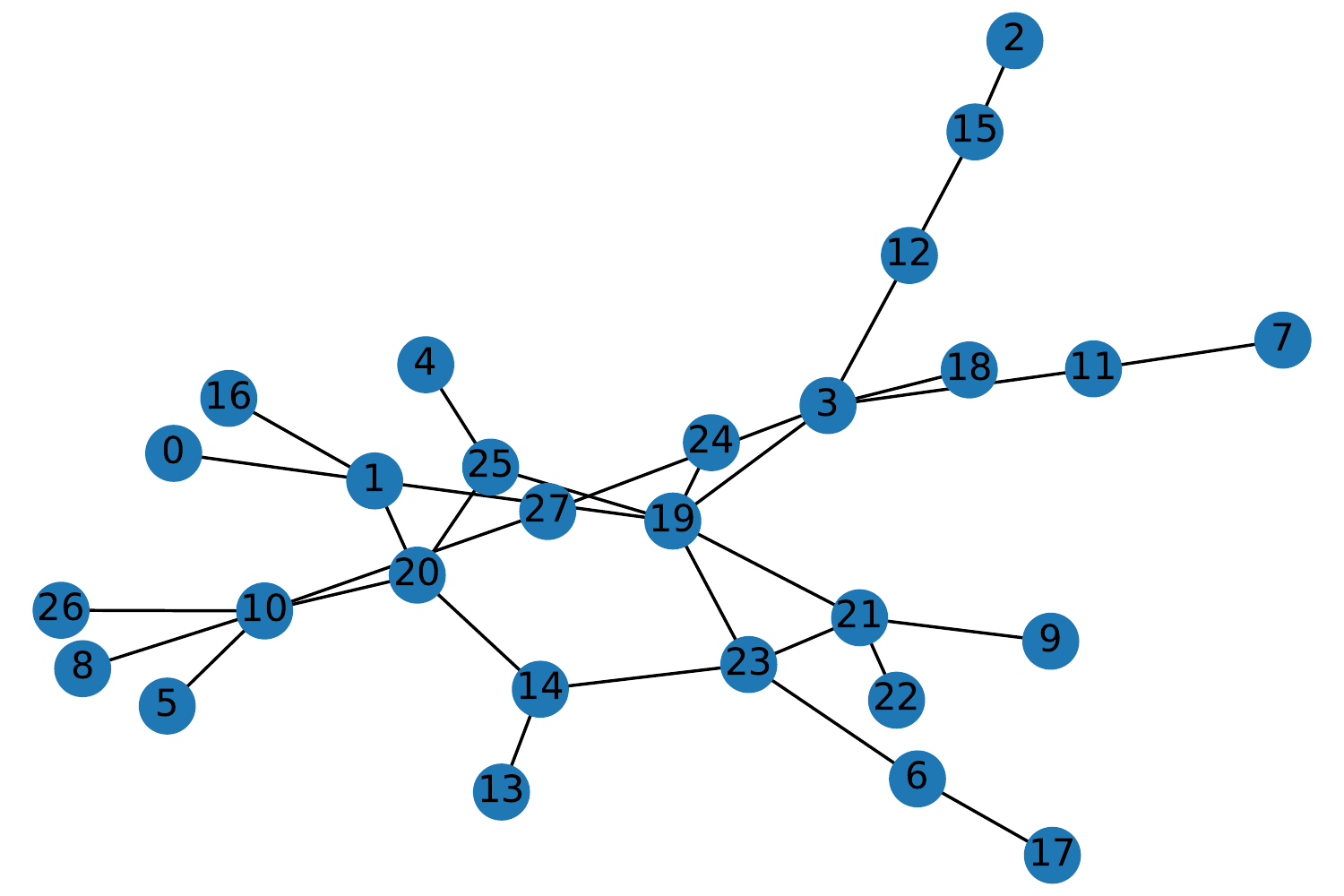}
\captionof{figure}{Sparser, chain-like architectures typically give a lower top one accuracy and so are discouraged by the controller network.}

\end{document}